\documentclass[11pt]{article}
\pdfoutput=1
\usepackage{jcappub}
\usepackage{slashed}
\usepackage{stackengine}
\usepackage{amsmath}
\usepackage{bm}
\usepackage[frak=euler,scr=boondox,bb= pazo]{mathalfa}

\usepackage[shortlabels]{enumitem}
\def\be{\begin{equation}}
\def\ee{\end{equation}}
\def\bea{\begin{eqnarray}}
\def\eea{\end{eqnarray}}
\usepackage[shortlabels]{enumitem}

\begin{document}

\title{Bio-inspired Machine Learning:\\
 programmed death and replication}  

\author[a,b]{Andrey Grabovsky}

\emailAdd{agrabovsky@gmail.com}

\author[c,d]{and Vitaly Vanchurin}

\emailAdd{vitaly.vanchurin@gmail.com}

\date{\today}

\affiliation[a]{Budker Institute of Nuclear Physics, 11, Lavrenteva avenue, 630090, Novosibirsk, Russia}

\affiliation[b]{Novosibirsk State University, 630090, 2, Pirogova street, Novosibirsk, Russia}

\affiliation[c]{National Center for Biotechnology Information, NIH, Bethesda, Maryland 20894, USA}

\affiliation[d]{Duluth Institute for Advanced Study, Duluth, Minnesota, 55804, USA}

\abstract{We analyze algorithmic and computational aspects of biological phenomena,  such as replication and programmed death, in the context of machine learning. We use two different measures of neuron efficiency to develop machine learning algorithms for adding neurons to the system (i.e. replication algorithm) and removing neurons from the system (i.e. programmed death algorithm). We argue that the programmed death algorithm can be used for compression of neural networks and the replication algorithm can be used for improving performance of the already trained neural networks. We also show that a combined algorithm of programmed death and replication can improve the learning efficiency of arbitrary machine learning systems. The computational advantages of the bio-inspired algorithms are demonstrated by training feedforward neural networks on the MNIST dataset of handwritten images. }
   
\maketitle

\section{Introduction}

Artificial neural networks \cite{Galushkin,Schmidhuber, Haykin} have been successfully used for solving computational problems in natural language processing, pattern recognition, data analysis etc. In addition to the empirical results a number of statistical approaches to learning where developed \cite{Hopfield, Vapnik, Bottleneck2} (see also \cite{Roberts} for a recent book on the subject) and some steps were taken towards developing a fully thermodynamic theory of learning \cite{learningtheory}. The thermodynamic theory was recently applied to model biological systems with evolutionary phenomena viewed as (either fundamental or emergent) learning algorithms \cite{biology1, biology2}. In particular, the so-called programmed death and replication (of information processing units, such as cells or individual organisms) were shown to be of fundamental importance for biological evolution modeled through learning dynamics. More generally, many statistical \cite{learningtheory}, quantum \cite{emergentquantum}, critical \cite{criticality}  and even gravitational \cite{quantumgravity} systems can be modeled using learning dynamics, and perhaps the entire universe may be viewed as a neural network that is undergoing learning evolution \cite{wnn}. However, for the quantum behavior to emerge it is essential that the total number of neurons is not fixed and could change over time \cite{emergentquantum}. This implies that in the emergent quantum systems, individual neurons should be constantly removed and added, similar to biological organisms, where individual cells are constantly die and replicate. In this paper, we analyze the machine learning algorithms that are inspired, first and foremost, by biology  \cite{biology1} (i.e. the programmed death and replication), but at the same time have direct connection to physics in general \cite{wnn} and to emergent quantum mechanics in particular \cite{emergentquantum} (i.e. removal and addition of neurons). 

Consider an artificial neural network that is being trained for some training dataset using some version of stochastic gradient descent. In this very general setup all of the neurons process information, but it is not clear which ones are more efficient and which ones are less efficient. For example, if we are to remove a single neuron what should it be so that the overall increase of the average loss function would be minimal? In other words, how do we find the least efficient neuron that has the least impact on the overall performance of the neural network? Can such a neuron be identified locally (e.g. by analyzing its state, bias and weights) or do we need to study global properties of the network (e.g. by analyzing non-local statistical or spectral properties )? If all neurons with low efficiency (or least loaded) can be identified, then one may be able to develop an algorithm (i.e. programmed death algorithm) that would be useful, for example, for compression of neural networks. Likewise, if one can identify the neurons with high efficiency (or most loaded), then, perhaps, one can use this information to develop replication algorithms where additional neurons would be added to the system to reduce the load on the most efficient neurons. Moreover, the two algorithms may also be used in conjunction (with programmed death followed by replication) in order to improve the learning rate of the existing machine learning algorithms. In this paper we will give answers to all of the above questions by carrying out analytical calculations (a more general, but only approximate method) and by conducting numerical experiments (a more special, but exact method). 

The paper is organized as follows. In the following section we discuss the basics of artificial neural network and of the learning dynamics.  In Sec. \ref{sec:efficiency} we perform statistical analysis of the learning dynamics and introduce two different definitions of neuron efficiency.  In Sec. \ref{sec:deathbirth} we develop, the ``programmed death'', the ``replication'' and the combined, i.e. programmed death followed by replication, algorithms. In Sec. \ref{sec:numerics} we present numerical results for feedforward neural networks with two hidden layers and different architectures. In Sec. \ref{sec:discussion} the main results are summarized and discussed.

\section{Neural networks}\label{sec:neuralnetwork}

A classical neural network with $N$ neurons can be defined as a septuple $({\bf x}, \hat{P}, p_\partial, \hat{w}, {\bf b}, {\bf f}, H)$, where
\begin{enumerate}
\item ${\bf x}$  is a (column) state vector of neurons,  
\item $\hat{P}$ is a boundary projection operator to subspace spanned by input/output neurons, 
\item $p_\partial(\hat{P} {\bf x})$ is a probability distribution which describes the training dataset, 
\item $\hat{w}$ is  a weight matrix,
\item ${\bf b}$ is a (column) bias vector, 
\item  ${\bf f}({\bf y})$ is an activation map and  
\item  $H({\bf x}, {\bf b}, \hat{w})$ is a loss function. 
  \end{enumerate}
The training data are associate only with boundary neurons  $\hat{P}{\bf x}(t)$ that are updated periodically from the probability distribution $p_\partial(\hat{P} {\bf x})$, but the period depends on the architecture. For example, for a feedforward architecture, the period could equal to the number of layers so that, in-between updates of the training data,  the signal has time to propagate throughout the entire network. In contrast to the boundary neurons, evolution of the bulk neurons depends on the state of all neurons,
\be
\left ( \hat{I} - \hat{P} \right ){\bf x}({t+1}) =\left ( \hat{I} - \hat{P}\right ) {\bf f} \left ( \hat{w} {\bf x}(t)+ {\bf b} \right),\label{eq:bulk_eom}
\ee
where the activation map acts separately on each component, i.e. $f_i ( {\bf y} ) = f_i(y_i) $.  For the sake of concreteness, we set activation functions on all boundary neurons to be linear
\be
\hat{P} {\bf f} ( {\bf y} ) =  \hat{P} {\bf y}
\ee
and on all bulk neurons to be hyperbolic tangent
\be
\left ( \hat{I} - \hat{P} \right ) {\bf f} ( {\bf y} ) =  \left ( \hat{I} - \hat{P} \right ) \tanh \left ( {\bf y} \right ).
\ee

The main objective of machine learning is to find bias vectors ${\bf b}$ and weight matrices $\hat{w}$ which minimize a time (or ensemble) average of some suitably defined loss  function. For example, the boundary loss function is given by 
\bea
H_\partial({\bf x}, {\bf b}, \hat{w}) &=&  \frac{1}{2} \left ( {\bf x}  - {\bf f} \left ( \hat{w} {\bf x}+ {\bf b} \right) \right )^T  \hat{P} \left (  {\bf x}  - {\bf f} \left ( \hat{w} {\bf x}+ {\bf b} \right) \right ) 
\label{eq:boundry_loss} 
\eea
where because of the inserted projection operator $\hat{P}$ the sum is taken over squared error at only boundary neurons. For example, in a feedforward architecture there is no error in the input boundary layer and all of the error is on the output boundary layer due to a mismatch between propagated data and training data.  Another example is the bulk loss function defined as
\bea
H({\bf x}, {\bf b}, \hat{w}) &=&  \frac{1}{2} \left ( {\bf x}  - {\bf f} \left ( \hat{w} {\bf x}+ {\bf b} \right) \right )^T \left (  {\bf x}  - {\bf f} \left ( \hat{w} {\bf x}+ {\bf b} \right) \right ) +  { V}({\bf x})\label{eq:bulk_loss} 
\eea
where in addition to the first term, which represents a sum of local errors over all neurons, there may be a second term which represents local objectives. 

During learning the trainable variables (i.e. bias vector ${\bf b}$ and weight matrix $\hat{w}$) are continuously adjusted (or transformed) 
\bea
b_i(t+T) &=& b_i(t)  - \gamma \frac{\partial \langle H({\bf x}, {\bf b}, \hat{w}) \rangle }{\partial b_i}\\
w_{ij}(t+T) &=& w_{ij}(t)  - \gamma \frac{\partial \langle H({\bf x}, {\bf b}, \hat{w}) \rangle}{\partial w_{ij}}
\eea
where the time-averaged quantities are defined as
\be
\left \langle ... \right \rangle \equiv \lim_{T\rightarrow \infty} \frac{1}{T} \sum_{t=1}^T ... \,\label{eq:mean}
\ee
and the time interval $T$ can depend on the mini-batch size and on the number of layers. Note that in general the ensemble average (for a given training dataset $p_\partial(\hat{P} {\bf x})$) and the time average (for a given time interval $T$) need not be the same, but, if the trainable variables (i.e. weights and biases) change very slowly, the two averages are approximately the same.  

In the long run, the learning system settles down in a local minimum of the average loss function and the learning effectively stops. For certain algorithms the system can still transition to an even lower minimum of the loss function, but such transitions are usually exponentially suppressed. The main problem is that in a local minimum continuous transformations (e.g. stochastic gradient decent) cannot be effective and discontinuous transformations must be performed instead. In Secs. \ref{sec:deathbirth} and \ref{sec:numerics} we shall describe one such transformation by combining two algorithms: programmed death and replication. More generally, the programmed death and replication transformations (or algorithms) need not be combined and can be used separately. The programmed death algorithm may be used to compress, either gradually (i.e. one neuron at a time) or suddenly (i.e. a bunch of neurons at once), the neural network. Such compression could be relevant, for example, if relatively large computational resources are available during the training phase, but the resources are limited during predicting phase. In addition, the replication algorithm may be used for improving the performance of an already pre-trained neural network. This can be done, for example, by adding new neurons in order to assist the most efficient neuron. 

\section{Statistical analysis}\label{sec:efficiency}

In the previous section, we mentioned the discrete machine learning algorithms (programmed death and replication) that rely on determining the efficiency of individual neurons. In this section we will describe the two definitions of neuron efficiency that will be used in Sec. \ref{sec:deathbirth} for developing these algorithms and in Sec. \ref{sec:numerics} for presenting the results of the numerical experiments.

\subsection{Covariance matrix}

To study the statistical efficiency of neurons it is convenient to introduce a covariance matrix,
\be
C_{ij}\equiv  \left \langle (x_i - \langle x_i \rangle )  (x_j - \langle x_j \rangle ) \right \rangle
\equiv  \left \langle \Delta x_i \Delta x_j \right \rangle
= \left \langle x_i x_j \right \rangle - \left \langle x_i \right \rangle  \left \langle x_j \right \rangle.
\ee
which is a positive definite symmetric matrix whose eigenvalues $\lambda_k$ are real nonnegative numbers and the corresponding orthonormal eigenvectors ${\bf v}^{(k)}$ are given by
\be
\hat{C} {\bf v}^{(k)} = \lambda_k {\bf v}^{(k)}.\label{eq:eigen}
\ee
If the states of neurons are approximately linearly dependent,
\be
\sum_{k}a_{k}x_{k}\approx a_0,\label{eq:dependence1}
\ee
then at least one of the eigenvalues must be zero. For example, if $\lambda_1=0$, then by setting 
 \be
v^{(1)}_k = a_{k} \label{eq:aj2}
\ee
we get
\be
\sum_k  v^{(1)}_k \langle x_k \rangle  =  \left \langle \sum_k a_k   x_k \right \rangle  \approx \langle a_0   \rangle =   a_0,\label{eq:a02}
 \ee
or 
\be
\sum_k \left (x_k -  \langle x_k \rangle \right )   v^{(1)}_k  \approx 0. \label{eq:apprx1}
\ee
in agreement with zero-eigenvalue equation
\begin{equation}
{\hat{C}\mathbf{v}^{(1)}=\sum_{k}\hat{C}_{ik}\mathbf{v}_{k}^{(1)}=\langle\Delta
x_{i}\sum_{k}a_{k}\Delta x_{k}\rangle\approx 0.}
\end{equation}
In this limit any one of the neurons can be removed after appropriate adjustment of weights. 

\subsection{Efficiency of neurons}

In general, $\lambda_i \neq 0$ for all $i$ and then one can define efficiency of individual neurons as the degree of non-linearity or how poorly the output of a given neuron can be approximated by a linear function of the outputs of all other neurons (for a given training dataset $p_\partial(\hat{P} {\bf x})$). The accuracy of approximation of the state of neuron $k$ is high (and thus the efficiency should be low) if $\lambda_i $ is small and $\left (v^{(i)}_k\right)^2$ is large. Therefore we can define the efficiency of neuron $k$ as
\be
E'_k = \min_i \frac{ \lambda_i}{\left (v^{(i)}_k\right)^2} ,\label{eq:efficiency1}
\ee
where the smallest ratio  is obtained by looking at all eigenvalues and the corresponding eigenvectors. Then we use the $i$-th equation \eqref{eq:apprx1}  to remove the $k$'th neuron. 
\begin{equation}
x_{k}=\frac{\sum_{j}\mathbf{v}_{j}^{(i)}x_{j}-\sum_{j\neq k}\mathbf{v}%
_{j}^{(i)}x_{j}}{\mathbf{v}_{k}^{(i)}}\approx\frac{\langle\sum_{j}%
\mathbf{v}_{j}^{(i)}x_{j}\rangle-\sum_{j\neq k}\mathbf{v}_{j}^{(i)}x_{j}%
}{\mathbf{v}_{k}^{(i)}}.
\end{equation}
By doing so we make the following mistake in $x_{k}$ 
\begin{equation}
\frac{\sum_{j}\mathbf{v}_{j}^{(i)}\Delta x_{j}}{\mathbf{v}%
_{k}^{(i)}}=O(\sqrt{E_{k}^{\prime}}),
\end{equation}
since%
\begin{equation}
\hat{C}_{ij}\mathbf{v}_{i}^{(m)}\mathbf{v}_{j}^{(n)}=\left \langle (\sum
_{i}\mathbf{v}_{i}^{(m)}\Delta x_{i} )(\sum_{j}\mathbf{v}_{j}^{(n)}\Delta
x_{j})\right \rangle=\lambda_{n}\delta^{mn}.
\end{equation}
Unfortunately, such algorithm may not be very useful in practice since it involves calculation of the covariance matrix --- a computational task which scales as ${\cal O}(N^2)$. What we really want is to be able to identify an approximate linear dependence of  neurons, but with an algorithm whose complexity would scale as ${\cal O}(N)$. 

Consider a linear expansion of activation function
\be
x_i = f_i \left (  y_i  \right ) +  f'_i \left ( y_i \right ) \sum_k w_{ik}  \left (  x_k - \langle x_k \rangle \right )+ ... 
\ee
where 
\be
y_i \equiv  \sum_k w_{ik}  \langle x_k \rangle + b_i.
\ee
At the zeroth order $ \left \langle x_i \right \rangle \approx f_i \left ( y_i \right )$ and at the first order
\be
x_i - \langle x_i \rangle  \approx  f'_i \left (  y_i \right )  \sum_k  w_{ik}  \left (  x_k - \langle x_k \rangle \right ) \label{eq:firstord}.
\ee
If the direct impact of neuron $k$ on neuron $i$ is small, then 
\be
C_{ii}  \gg  f'_i \left ( y_i \right )^2 w_{ik}^2  C_{kk} \label{eq:least}
\ee
or
\be
1  \gg   f'_i \left ( y_i \right )^2 w_{ik}^2  C_{kk}  C_{ii}^{-1}. \label{eq:least2}
\ee
In this limit all possible variations of the signal 
\be
x_k \approx \langle x_k \rangle \label{eq:apprx2}
\ee 
do not significantly modify the expected signal $x_i$, and (if our task is to implement a programmed death algorithm) then we must identify the least efficient neuron by summing over all impacts that a given neuron $k$ has on all other neurons $i$, i.e.
\be
E_k \equiv C_{kk}  \sum_i f'_i \left ( y_i \right )^2 w_{ik}^2  C_{ii}^{-1}. \label{eq:efficiency2}
\ee
 If $E_k\ll 1$, then we can drop all of the connection from neuron $k$ to neuron $i$, or the neuron $k$ can be removed without sacrificing much of the neural network performance. More precisely, we can set $w_{ik}=0$ for all $i$ and then use \eqref{eq:apprx2} to readjust biases $b_i$'s of all other neurons so that the input to $i$-th neuron remains approximately the same. See Eq. \eqref{eq:input} with $k=1$. Note that all of $C_{jj}$'s and all of $f'_j \left ( y_j \right )$'s can be calculated locally by analyzing statistics of the signals for each of the neurons separately --- a computational task which scales as ${\cal O}(N)$. 

\subsection{Conditional distribution} 

In the limit opposite to \eqref{eq:least} or \eqref{eq:least2}, the impact of neuron $k$ on neuron $i$ is large,
\be
C_{ii}  \ll  f'_i \left ( y_i \right )^2 w_{ik}^2  C_{kk} \label{eq:most}
\ee
or
\be
1  \ll  f'_i \left ( y_i \right )^2 w_{ik}^2  C_{kk}  C_{ii}^{-1} \label{eq:most2},
\ee
and variations of signal $x_k$ can significantly modify the signal $x_i$. However, since variations of $ f'_i \left (  y_i \right ) w_{ik}  \left (  x_k - \langle x_k \rangle \right ) $ must remain much larger than variations of  $ \left (  x_i - \langle x_i \rangle \right ) $, variations of all other incoming signals $f'_i \left (  y_i \right ) w_{ij} \left (  x_j - \langle x_j \rangle \right ) $ (where $j\neq k$) must be anti-correlated with $f'_i \left (  y_i \right ) w_{ik} \left (  x_k - \langle x_k \rangle \right ) $.  More precisely, all of the incoming signals must be approximately linearly dependent and then the output $x_k$ can be approximated as a linear function of the outputs of all other neurons
\be
 x_k \approx    \sum_{j} \frac{w_{ij}}{w_{ik} } \langle x_j \rangle -  \sum_{j\neq k} \frac{w_{ij}}{w_{ik} }  x_j +   \frac{x_i - \langle x_i \rangle }{f'_i \left (  y_i \right ) w_{ik} }    \label{eq:xk_aprx}.
\ee
Then the conditional distribution for variable $x_k$ (with all other $x_j$'s for $j\neq k$ fixed) can be modeled as a Gaussian
\be
p(x_k) \propto \exp \left ( - \frac{1}{2} \frac{\left( \mu_i - x_k\right )^2}{\sigma^2_i} \right ) \label{eq:prob}
\ee
with mean 
\be
\mu_i \equiv  \sum_{j} \frac{w_{ij}}{w_{ik} } \langle x_j \rangle -  \sum_{j\neq k} \frac{w_{ij}}{w_{ik} }  x_j 
\ee
and variance
\be
\sigma_i^2  = \left( f'_i \left (  y_i \right )^2 w_{ik}^2 C_{ii}^{-1} \right )^{-1}.
\ee

To improve the estimate further, we can average over all such approximations for $i\neq k$ and then the overall distribution is proportional to a product of Gaussians 
\be
p(x_k) \propto \exp \left ( - \frac{1}{2} \sum_{i \neq k} \frac{\left( \mu_i - x_k\right )^2}{\sigma^2_i} \right ) \propto  \exp \left ( - \frac{1}{2} \frac{\left( M_k - x_k\right )^2}{S_k^2}  \right )
\ee
with mean 
\be
M_k = \frac{\sum_{i \neq k}  \mu_i \sigma_i^{-2}}{\sum_{i \neq k}  \sigma_i^{-2}} =   S_k^2  \sum_{i \neq k} f'_i \left (  y_i \right )^2 w_{ik} C_{ii}^{-1}   \left (\sum_{j} w_{ij} \langle x_j \rangle -  \sum_{j\neq k} w_{ij}  x_j  \right )\label{eq:apprx3}
\ee
and variance 
\be
S_k^2 =  \left ( \sum_{i \neq k}  \sigma_i^{-2} \right )^{-1} =  \left ( \sum_{i \neq k} f'_i \left (  y_i \right )^2  w^2_{ik} C_{ii}^{-1}  \right )^{-1}. \label{eq:variance}
\ee
Note that the neurons efficiency \eqref{eq:efficiency2} is inversely proportional to the variance 
\be
E_k = \frac{C_{kk}}{S_k^2}
\ee
and so the variance is large for neurons with smaller efficiencies and vise versa. 

\section{Machine learning algorithms} \label{sec:deathbirth}

In this section we develop the three machine algorithms: programmed death, replication and combined (i.e. programmed death followed by replication), using the two measures  of efficiencies of individual neurons, or just efficiencies, that were introduce in the previous section. Without loss of generality, we assume that the least efficient neuron is the $k=1$ neuron where the efficiency is defined either by Eq. \eqref{eq:efficiency1} (and then the linear dependence of Eq. \eqref{eq:apprx1} should be used) or by Eq.\eqref{eq:efficiency2} (and then the linear dependence of Eqs. \eqref{eq:apprx2} or \eqref{eq:apprx3} should be used).   For the numerical experiments of Sec. \ref{sec:numerics} we shall refer to
\begin{enumerate}[{A}1.] 
\item ``Connection cut'' algorithm ---  method of Eqs. \eqref{eq:apprx2} and \eqref{eq:efficiency2}
\item ``Probability'' algorithm --- Eqs. \eqref{eq:apprx3} and \eqref{eq:efficiency2}.
\item ``Covariance'' algorithm --- method of Eqs. \eqref{eq:apprx1} and \eqref{eq:efficiency1}
\end{enumerate}
However, once the least efficient $k=1$ neuron is identified (using either \eqref{eq:efficiency1} or \eqref{eq:efficiency2}) and once an approximate linear relation is established (using either \eqref{eq:apprx1}, \eqref{eq:apprx2} or \eqref{eq:apprx3})
\be
\sum_{j}    a_j x_j   \approx  a_0 \label{eq:dependance}
\ee
all three algorithms (i.e. covariance, connection cut and probability) are treated similarly. Also note, that in a feedforward architecture the approximate linear dependence \eqref{eq:dependance} can only be established between neurons $j$ on the same layer with the least efficient neuron $k=1$.

\subsection{Programmed death}

In the programmed death algorithm the least efficient neuron is removed from the neural network and the rest of the network is re-wired to accommodate the changes.  It is important to emphasize that the removed neuron must be in the bulk (i.e. in the hidden layers) so that no training data are associated with it directly. 

The activation dynamics of neuron $x_i = f_i(y_i)$ is determined by the state of all other neurons $x_j$ only through a linear function 
\be
y_i = \sum_j w_{ij} x_j + b_i
\ee
which can be approximated using  \eqref{eq:dependance} as
\bea
y_i &= &   \sum_{j\neq 1} w_{ij} x_j + w_{i1}x_1  + b_i \notag \\
 &\approx&   \sum_{j\neq 1} w_{ij} x_j + w_{i1} \left (\frac{a_0}{a_1}  - \sum_{l\neq 1} \frac{a_l}{a_1} x_l  \right ) +b_i \notag \\
 &=&   \sum_{j\neq 1} \left ( w_{ij} - w_{i1} \frac{a_j}{a_1} \right ) x_j + \left (b_i + w_{i1} \frac{a_0}{a_1}  \right ).\label{eq:input}
 \eea
 If we are to disconnect a neuron from the network with a minimal modification to the states of other neurons, then we have to readjust the biases and weights so that $y_i$'s remain approximately the same. This can be done using the following discrete transformation of the weight matrix 
\bea
w'_{ij} = \begin{cases}  0  &\;\;\;\text{if}\;\;\; i=1\;\;\text{or}\;\;j=1\\
 w_{ij}  - w_{i1} \frac{a_j}{a_1}  &\;\;\;\text{otherwise}
\end{cases}
\eea
and of the bias vector
\bea
b'_{i} = \begin{cases}  0  &\;\;\;\text{if}\;\;\; i=1\\
b_{i}  + w_{i1} \frac{a_0}{a_1}    &\;\;\;\text{otherwise}.
\end{cases}
\eea
In other words, the transformation sets all of the signals to and from the ``dead'' neuron to zero and readjusts all other weights and biases so that equations \eqref{eq:input} are approximately satisfied for $i\neq 1$. In this way, it is ensured that the performance of the neural network is not significantly altered or that the value of the average loss function, for a given set of training data, is not significantly changed. 

The programmed death algorithm is equivalent to a biological phenomenon known as the programmed death, e.g. programmed cell death. From a more practical point of view, the algorithm can be used for compression of neural networks for further use, for example, on devices with constrained computational resources. 

\subsection{Replication}

In the replication algorithm the learning system adds a new neuron to the neural network in order to reduce the load on the most efficient neuron, not necessarily immediately, but in the long run. Once again, the efficiency of individual neurons can be define by either \eqref{eq:efficiency1} or \eqref{eq:efficiency2}, but now the main challenge is to introduce coupling between the two neurons: the new (or ``child'') neuron ``$c$'' and the most efficient (or ``parent'') neuron ``$p$''. By following the biological analogy of the phenomenon of cell replication, we shall only couple the neurons at the time of the replication. This must correspond to some clever reinitialization of biases and weights to and from the child and parent neurons that in the long run would lead to the largest decrease of the average loss function.

For example, consider reinitialization described by a discrete transformation of the weight matrix
\bea
w'_{ij} = \begin{cases}  w_{pj}  &\;\;\;\text{if}\;\;\; i=c\\
\frac{1}{2} w_{ip}  &\;\;\;\text{if}\;\;\; j=c\;\;\text{or}\;\;j=p \\
 w_{ij}  &\;\;\;\text{otherwise}
\end{cases}\label{eq:equal split}
\eea
and of the bias vector
\bea
b'_{i} = \begin{cases}   b_p &\;\;\;\text{if}\;\;\; i=c \\
b_i &\;\;\;\text{otherwise}.
\end{cases}
\eea
This transformation first splits in half all of the outgoing  weights from the parent neuron ``$p$'' and then copies all of the outgoing weights from the parent  ``$p$'' to child neuron ``$c$''. As a result, the overall performance of the neural network is not altered and we end up with two identical (or replicated) neurons that are, however, linearly dependent 
\be
x_c = x_p. 
\ee
This means that if the programmed death procedure were to be executed right after replication, then it would immediately identify and delete one of these neurons. Moreover, if the two neurons carry exactly the same information, the positive effect of the replication on learning would be only marginal.

To avoid the problem of immediate removal of a newly replicated neuron and to improve the learning efficiency we can set the outgoing weights from the child and parent neurons to be arbitrary given that 
\be
w'_{i p} +w'_{i c} = w_{ip}.
\ee 
For example, we can split the outgoing signals between the child and parent neurons by defining a  discrete transformation of the weight matrix
\bea
w'_{i p} &=& \chi_i  w_{ip} \label{eq:weight splitting1}\\
w'_{i c} &=& (1- \chi_i)  w_{ip} \label{eq:weight splitting2}
\eea
where $\chi_i \in \{0, 1\}$ is a random bit. Note, that the randomization procedure can be improved further by employing continuous probability distributions $P(\chi_i)$ which may or may not be symmetric.

The replication algorithm can be used for increasing the effective dimensionality of the space of trainable variables in the regions where the learning resources are most needed. This may be important for initialization of the neural networks as well as for improving the performance of already trained neural networks.  More generally, as we shall see below, the replication algorithm can be used in conjunction with programmed death algorithm for improving the convergence of arbitrary learning systems.

\section{Numerical results}\label{sec:numerics}

In the previous section we developed two bio-inspired machine learning algorithms (i.e. programmed death, replication) and suggested that the programmed death followed by replication can be used for increasing the learning efficiency of arbitrary algorithms. In this section we will analyze the performance of these algorithms by training a feedforward neural network with four layers (i.e. two hidden layers) and different neural architectures:
\begin{enumerate}[{N}1.] 
\item $784$ neurons $\rightarrow$ $5$ neurons $\rightarrow$ $20$ neurons $\rightarrow$ $10$ neurons 
\item $784$ neurons $\rightarrow$ $10$ neurons $\rightarrow$  $10$ neurons $\rightarrow$ $10$ neurons 
\item  $784$ neurons $\rightarrow$ $20$ neurons $\rightarrow$ $5$ neurons $\rightarrow$ $10$ neurons 
\item  $784$ neurons $\rightarrow$ $100$ neurons $\rightarrow$ $5$ neurons $\rightarrow$ $10$ neurons 
\end{enumerate}
The neural networks were trained on the MNIST \cite{MNIST} dataset of 60000 handwritten images, $p_\partial(\hat{P} {\bf x})$; with linear activation function, $f(y)=y$, for boundary neurons (i.e. first and fourth layers); with non-linear activation function, $f(y)=\tanh(y)$, for bulk neurons (i.e. second and third layers); and with cross-entropy loss function, $H({\bf x}, {\bf b}, \hat{w})$. The training was done via the stochastic gradient descent method with the batch size $600$, momentum $0$, L2 regularization parameter $0.001$, and the constant learning rate $0.001$. For plotting the numerical results we use the neuron efficiency calculated according to Eq. \eqref{eq:efficiency1}  for the ``covariance'' algorithm, A3, and according to Eq. \eqref{eq:efficiency2} for the ``connection cut'' algorithm, A1, and the ``probability'' algorithm, A2 (see Sec. \ref{sec:deathbirth} for details). 

\subsection{Programmed death}

For numerical testing of the programmed death algorithms, described in the previous section, we trained the feedforward neural networks  N1, N2 and N3 for $50000$ epochs. 

\begin{figure}[h]
    \centering
    \includegraphics[width=\textwidth]{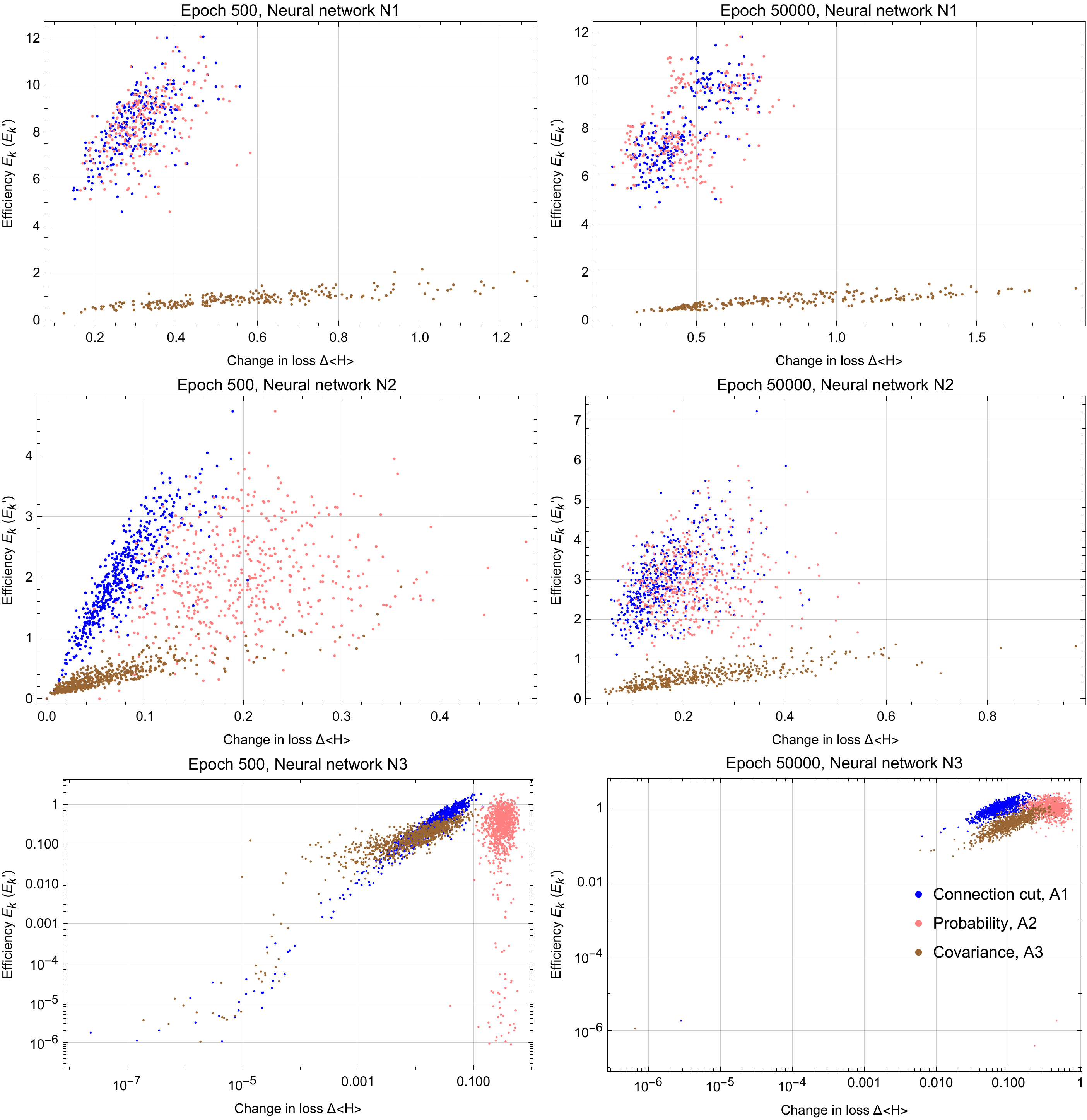}
    \caption{Efficiency of neuron, $E_k$ or $E_k'$, vs. change in the average loss function, $\Delta \langle H \rangle$, for algorithms A1 (first row), A2 (second row) and A3 (third row), and for neural architectures N1 (blue dots), N2 (pink dots) and N3 (brown dots) after 500 (left plots) and 50000 (right plots) epochs.}
    \label{fig:deltalossVsEfficiency}
\end{figure}

On Fig. \ref{fig:deltalossVsEfficiency}
 we plot the efficiency of neuron, $E_k$ or $E_k'$, vs. change in the average loss function, $\Delta \langle H \rangle$, (i.e. loss after neuron is removed minus loss before neuron is removed) for three different neural networks (N1, N2 and N3) and three different algorithms (A1, A2 and A3) at epochs 500 and 50000. For the individual runs,  efficiency of every neuron on the second layer is calculated, then each neuron is removed and the change in the loss function is calculated. Statistics is acquired by running fifty simulations with different initialisation for every algorithm and neural architecture. In the third row (or for the neural network N3) and for "connection cut" and "probability" plots (or for A1 and A3 algorithms) the log-log plot is used to show that there are many neurons with efficiency smaller than $\leq10^{-3}$ at epoch 500, but only one such neuron at epoch 50000. In programmed death algorithms, such low-efficiency neurons can be removed without significantly changing the performance of the neural network. 

\begin{figure}[h]
    \centering
    \includegraphics[width=\textwidth]{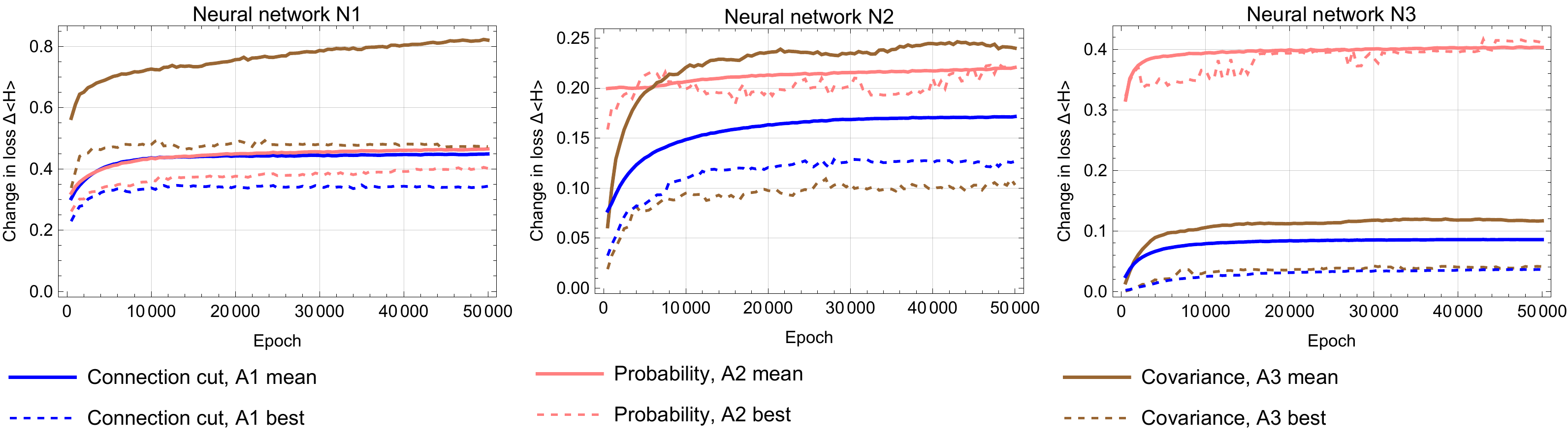}
    \caption{ $\Delta \langle H \rangle$ as a function of learning time for algorithms A1 (blue lines), A2 (pink lines) and A3 (brown lines) and for neural networks N1 (left plot), N2 (middle plot) and N3 (right plot).}
    \label{fig:average and best delta loss}
\end{figure}

On Fig. \ref{fig:average and best delta loss} \begin{figure}[h]
    \centering
     \includegraphics[width=\textwidth]{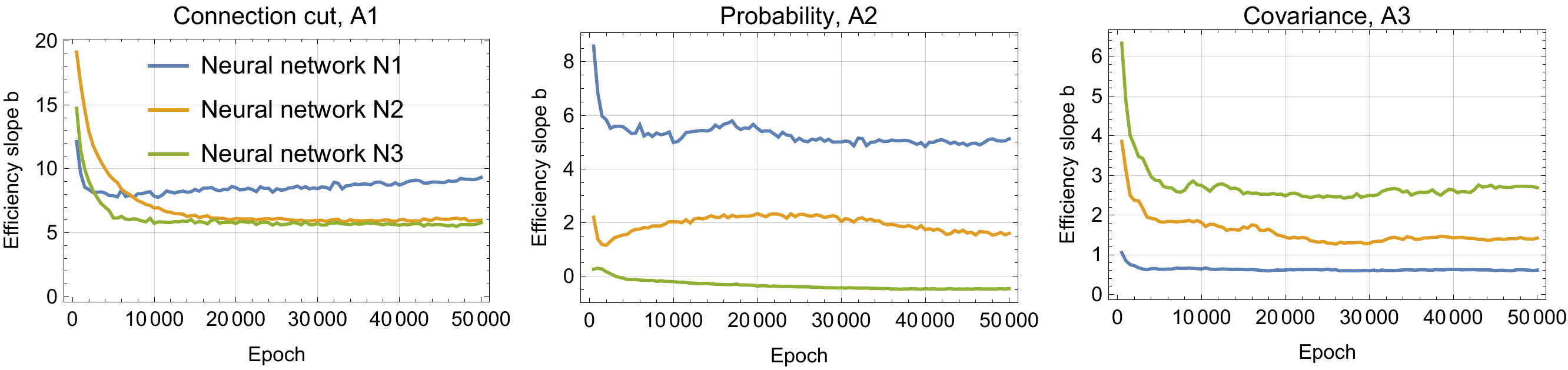}
    \caption{Slopes $b(t)$ of the linear fits (of $E_k$ (or $E_k'$) vs. $\Delta\langle H \rangle$) as a function of time $t$.}
    \label{fig: Efficiency slope}
\end{figure}we plot change in the average loss function, $\Delta \langle H \rangle$, as a function of time (or the number of epochs) for three different algorithms (A1, A2 and A3) and for three different neural architectures (N1, N2 and N3). For each run of the simulation every neuron on the second layer is removed and then $\Delta \langle H \rangle$ is calculated. For N1 neural architecture we executed 50 separate runs of the simulation, for N2 neural architecture --- 25 runs, and for N3 neural architecture --- 13 runs all with different initial conditions. The solid lines show the mean $\Delta \langle H \rangle$ averaged over all neurons on the second layer and all runs, and the dashed lines show the mean $\Delta \langle H \rangle$ for only the least efficient neuron on the second layer in each run averaged over all runs. The least efficient neuron is the one with the smallest efficiency, i.e. smallest $E_k'$ for the A1 and smallest $E_k$ for A2 or A3 algorithms. Clearly, only when the dotted line is much lower than the sold line (of the same color) the removal of the least efficient neuron would lead to the smallest distortion to the overall performance of the network. With this respect only for the neural networks N2 and N3, and only for the algorithm A2 (or pink lines) the corresponding probability method is not very useful. 

Next, we make linear fits of $E_k$ vs. $\Delta\langle H \rangle$ (for algorithms A1 and A2),
\be
E_k = a + b\,\Delta\langle H \rangle,
\ee
or $E_k'$  vs. $\Delta\langle H \rangle$ (for algorithm A3),
\be
E_k' = a + b\,\Delta\langle H \rangle,
\ee
for all times. On Fig. \ref{fig: Efficiency slope} we plot the slopes $b(t)$ as a function of learning time, where the data are obtained from the same runs as for Fig. \ref{fig:average and best delta loss}. Note that for algorithm A2 (second plot) and neural network N3 (green line) the slopes are almost zero (or slightly negative) and so the probability method is not very useful for identifying and removing a neuron which would give the smallest change in loss. In fact, according to Fig. \ref{fig:average and best delta loss}, one should remove the least efficient neuron using algorithm A1 or A2 for neural network N1 and using algorithm A1 or A3 for neural networks N2 and N3. Moreover, according to Fig. \ref{fig: Efficiency slope}, the algorithms A1 and A3 should work for all neural networks since the blue and yellow lines remain positive, which allows us to predict the effect of removing a given neuron, but the algorithm A2 would not work for neural network N3 since the green line on the rightmost plot remains near zero.

\subsection{Replication}

For numerical testing of the replication algorithms, described in the previous section, we trained the feedforward neural networks N1, N2 and N3 for $25000$ epochs. Then we use one of the algorithms to add one more neuron and run the simulation for  $10$, $100$, and $1000$ additional epochs.

\begin{figure}[h]
    \centering
    \includegraphics[width=\textwidth]{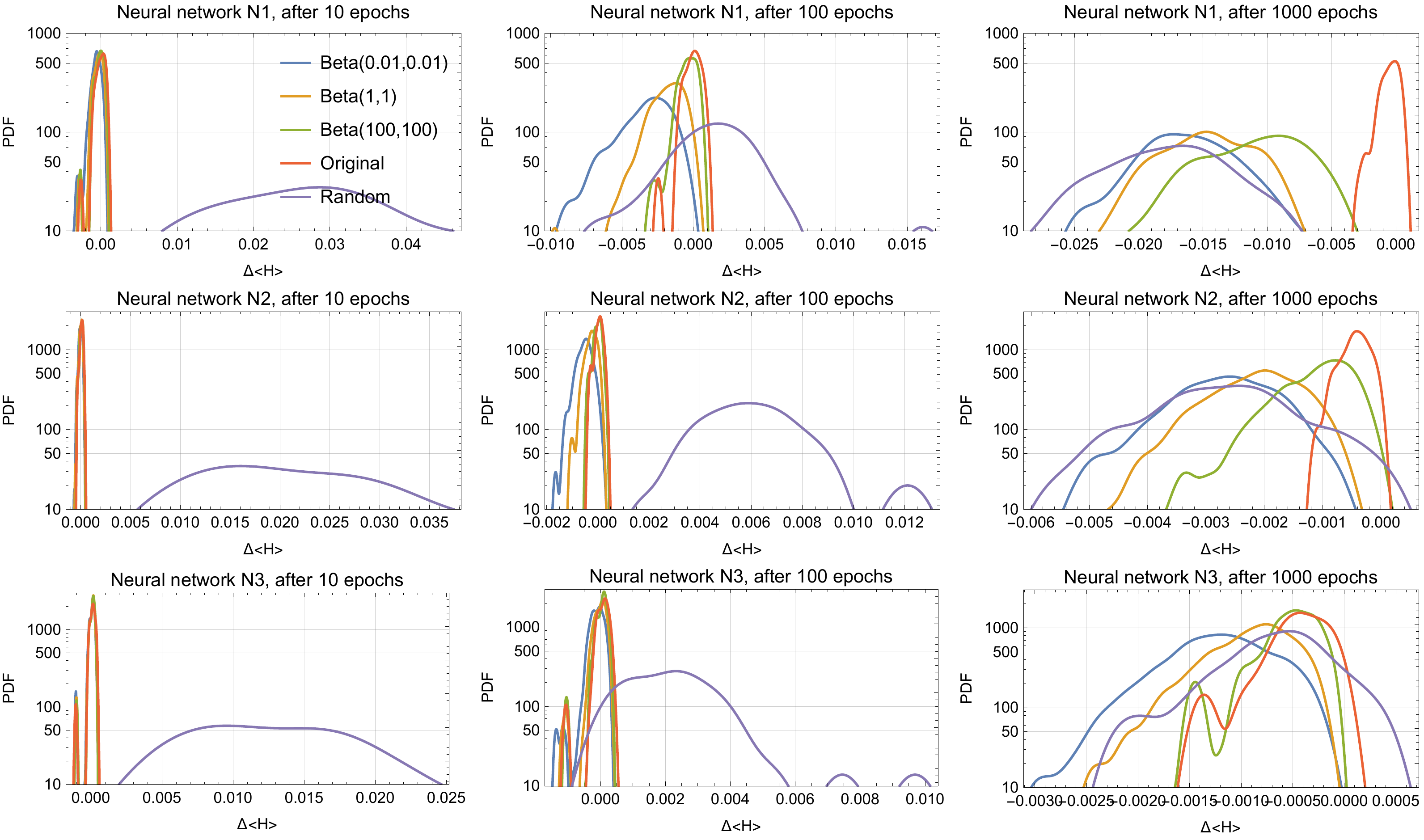}
    \caption{PDF  of $\Delta \langle H \rangle$ for neural networks N1 (first row), N2 (second row) and N3 (third row) after $10$ (first column), $100$ (second column), and $1000$ (third column) additional epochs. }
    \label{fig:add-neuron-PDF-graphs-for-25000-epoch-after-10-100-1000-epochs}
\end{figure}

On Fig. \ref{fig:add-neuron-PDF-graphs-for-25000-epoch-after-10-100-1000-epochs}
we plot PDF (or probability distribution function) of $\Delta \langle H \rangle$, acquired from fifty runs with different initialisation, for different neural networks and for different numbers of additional epochs. We use the $\beta$ distribution to model probability distribution $P(\chi_i)$ for splitting weights between the parent and child neurons in Eqs. \eqref{eq:weight splitting1} and \eqref{eq:weight splitting2}. The five lines on each plot describe five different algorithms:
\begin{enumerate}[{B}1.] 
\item ``Beta$(0.01,0.01)$'' --- neuron is replicated with Beta distribution and both shape parameters equal to $0.01$ (blue lines on Fig. \ref{fig:add-neuron-PDF-graphs-for-25000-epoch-after-10-100-1000-epochs}),
\item ``Beta$(1,1)$'' --- neuron is replicated with Beta distribution and both shape parameters equal to $1$ (yellow lines on Fig. \ref{fig:add-neuron-PDF-graphs-for-25000-epoch-after-10-100-1000-epochs}),
\item ``Beta$(100,100)$''--- neuron is replicated with Beta distribution and both shape parameters equal to $100$ (green lines on Fig. \ref{fig:add-neuron-PDF-graphs-for-25000-epoch-after-10-100-1000-epochs}),
\item ``Random''-- neuron with random weights (drawn from the same distribution as other weights) is added
(purple line),
\item ``Original''--- no new neurons are added (red lines on Fig. \ref{fig:add-neuron-PDF-graphs-for-25000-epoch-after-10-100-1000-epochs}).
\end{enumerate}
For algorithms B1, B2, and B3 the most efficient neuron (smallest $E_k'$ or $E_k'$) on the second layer  was replicated. One can see on Fig. \ref{fig:add-neuron-PDF-graphs-for-25000-epoch-after-10-100-1000-epochs} that after $1000$ additional epochs the most efficient algorithm is B1 (i. e. outgoing connections from a parent neuron are randomly split between parent and child neurons) and the least efficient algorithm is B3 (i. e. new outgoing weights from parent and child neuron equal to half of old outgoing weight from parent neuron). The main reason is that algorithm B1 creates a child neuron which is maximally independent from parent neuron while for algorithm B3 the parent and child neurons are equivalent  and it would take time for them to diverge due to stochastic learning dynamics. B4 algorithm is inefficient in the short run (i.e. after $10$ or $100$ epochs), but the algorithm becomes as efficient as B1 in a long run (i.e. after $1000$ epochs). 

To demonstrate the computational advantage of a combined algorithm, i.e.  programmed death followed by replication, we used algorithm A1 (for neuron removal) and B1 (for neuron addition) with neural architecture N4. The main reason for using N4 (as opposed to N1, N2 or N3) is that one needs a large number of neurons on the second layer for the effect to be most noticeable. We first run the simulation for $\Delta t$ epochs, calculate efficiencies (\ref{eq:efficiency2}) of all neurons on the second layer and use algorithm A1 to remove all neurons (but not more than $N/2$) whose efficiencies are less than a cutoff $\epsilon$. If $n$ neurons were removed, then we use algorithm B1 to replicates $n$ most efficient neurons and continue the simulation for another $\Delta t$ epochs and then execute the combined A1-B1 algorithm again, etc.

\begin{figure}[h]
    \centering
    \includegraphics[width=\textwidth]{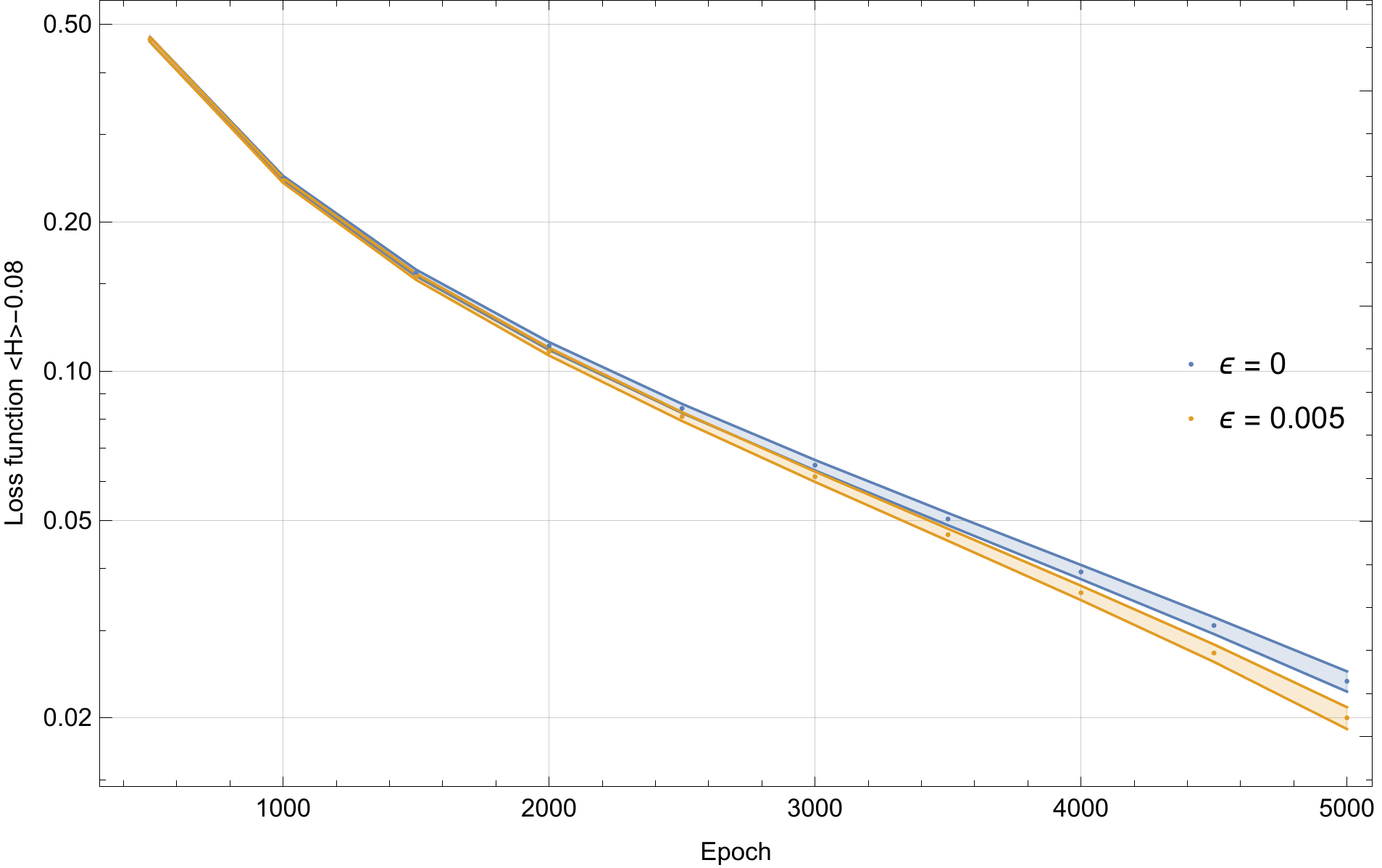}
    \caption{Log-linear plot of average loss function (minus asymptotic loss function) vs. time for a combined A1-B1 algorithm with cutoff $\epsilon = 0.005$ and $\Delta t=500$ epochs (yellow line with error bands) and for original algorithm $\epsilon = 0$ (blue line with error bands) averaged over 28 runs with different initialization.}
    \label{fig:cut-and-add}
\end{figure}

On Fig. \ref{fig:cut-and-add} we plot the average loss function for the combined A1-B1 algorithm for 5000 training epochs with $\epsilon = 0.005$ and $\Delta t=500$ epochs, and the original algorithm (or more precisely with $\epsilon = 0$ and $\Delta t=500$). The parameters $\epsilon=0.005$ and $\Delta t=500$ were chosen so that about $30\%$ of neurons would be affected by the algorithm after $\Delta t=500$ epochs, but this fraction goes to $15\%$ after $2 \Delta t=1000$ epochs, etc. The computational advantage of the combined A1-B1 algorithm is easy to understand. Indeed, the inactive neurons are constantly removed by the programmed death algorithm A1 and become active by the replication algorithm B1 which improves the learning efficiency for a carefully chosen cutoff threshold.

\section{Discussion}\label{sec:discussion}

In this article, we took a small step towards developing machine learning algorithms inspired by the well-known biological phenomena. In particular, we analysed the computational advantage of the programmed death and replication, since they represent the most essential phenomena not only in biology \cite{biology1}, but also in physics \cite{emergentquantum}. Indeed, in both biological and physical systems, the fundamental information processing units can either be added to the system or removed from the system, which gives rise to the biological phenomena of replication and programmed death \cite{biology1} and to the physical phenomena of emergent quantumness \cite{emergentquantum}. 

The developed programmed death and replication algorithms may have a wide range of applications in machine learning. For example, the programmed death algorithm may be useful for compression of neural networks for the use on devices with limited computational resources. In contrast, the replication algorithm may be useful for improving the performance of already trained neural networks on the devices where additional computational resources are available. We have also shown that a combination of programmed death and replication algorithms (i.e. reconnecting the least efficient neuron to reduce the load on the most efficient neuron) may be useful for improving the learning efficiency of an arbitrary machine learning system. 

More generally, when the machine learning system is stuck in a local minimum of the average loss function, the continuous transformations (e.g. stochastic gradient decent) cannot be efficient and a discrete transformation must be performed instead. With this respect the programmed death followed by replication is an example of such transformation.  Indeed,  the rewiring of connections from the least efficient (i.e. programmed death) to assist the most efficient neurons (i.e. replication) is a discrete transformation that may turn out to be useful for certain machine learning tasks. 

Although our analytical results are robust, the numerical results are only preliminary and a lot more numerical testing is needed in order to confirm the computational advantages of the proposed algorithms. Moreover, so far we have analyzed the discrete transformations that correspond to only two biological phenomena, programmed death and replication, and there are many other important biological \cite{biology1} and physical \cite{wnn} phenomena that can be analyzed in a similar manner which we leave for future work.  

{\it Acknowledgments.}  V.V. was supported in part by the Foundational Questions Institute (FQXi) and the Oak Ridge Institute for Science and Education (ORISE).


\begin{thebibliography}{10}



\bibitem{Galushkin} A.I. Galushkin, ``Neural Networks Theory,'' Springer, 396 p., (2007)

\bibitem{Schmidhuber}  J.~Schmidhuber, ``Deep Learning in Neural Networks: An Overview,'' Neural Networks. 61: 85-117. (2015)

\bibitem{Haykin} Haykin, Simon S. ``Neural Networks: A Comprehensive Foundation,'' Prentice Hall. (1999)

\bibitem{Vapnik}
 Vapnik, Vladimir N, ``The Nature of Statistical Learning Theory,'' Information Science and Statistics (2000)

\bibitem{Hopfield} J. J. Hopfield, "Neural networks and physical systems with emergent collective computational abilities", PNAS 79(8) pp. 2554-2558, 1982

\bibitem{Bottleneck2}
R.~ Shwartz-Ziv, N.~Tishby, ``Opening the black box of deep neural networks via information,'' arXiv:1703.00810 [cs.LG], (2017)


\bibitem{Roberts} 
Roberts, D., Yaida, S., Hanin, B. ``The Principles of Deep Learning Theory: An Effective Theory Approach to Understanding Neural Networks,'' Cambridge: Cambridge University Press. (2022)



\bibitem{learningtheory}
V. Vanchurin, ``Toward a theory of machine learning,''   Mach. Learn.: Sci. Technol. 2 035012 (2021)


\bibitem{biology1}
V. Vanchurin, Y. I. Wolf, M. O. Katsnelson, E. V. Koonin, ``Towards a theory of evolution as multilevel learning,'' Proc. Natl. Acad. Sci. U.S.A. 119 (2022)


\bibitem{biology2}
V. Vanchurin, Y. I. Wolf, E. V. Koonin, M. O. Katsnelson,  ``Thermodynamics of evolution and the origin of life,'' Proc. Natl. Acad. Sci. U.S.A. 119 (2022)


\bibitem{emergentquantum}
Mikhail I. Katsnelson, Vitaly Vanchurin, ''Emergent Quantumness in Neural Networks'', Foundations of Physics 51 (5):1-20 (2021), 

\bibitem{criticality} M.I. Katsnelson, V. Vanchurin, T. Westerhout, ``Self-organized criticality in Neural Networks,'' arXiv:2107.03402


\bibitem{quantumgravity}
V. Vanchurin,  ``Towards a Theory of Quantum Gravity from Neural Networks,'' Entropy, 24, 7 (2022)

\bibitem{wnn}
V. Vanchurin, ``The world as a neural network,'' Entropy, 22, 1210 (2020)



\bibitem{MNIST}
Y. LeCun, L. Bottou, Y. Bengio, and P. Haffner, ``Gradient-based learning applied to document
recognition,'' Proceedings of the IEEE, 86(11):2278-2324, 1998.


\end{thebibliography}
\end{document}